\documentclass[runningheads]{llncs}
\usepackage{graphicx}
\usepackage[latin1]{inputenc}
\usepackage{amssymb,amsmath,array}
\usepackage{enumitem}
\usepackage{multirow}
\usepackage{times}
\usepackage{xcolor}
\usepackage{xspace}
\usepackage{array}
\usepackage{booktabs}
\usepackage[symbol]{footmisc}

\definecolor{chromeyellow}{rgb}{1.0, 0.65, 0.0}

\begin{document}
\title{Real-Time Event Detection with Random Forests and Temporal Convolutional Networks \\for More Sustainable Petroleum Industry}

\author{Yuanwei Qu\inst{1} \thanks{Corresponding author} \and
Baifan Zhou\inst{2,1} \and
Arild Waaler\inst{1} \and
David Cameron\inst{1}}

\authorrunning{Qu et al.}
\titlerunning{Real-Time Event Detection with RF and TCN for Sustainablility}
%
\institute{
$^1$ University of Oslo, Oslo, Norway \\
\email{\{quy,baifanz,arild,davidbc\}@ifi.uio.no}\\
$^2$ Oslo Metropolitan University, Oslo, Norway}



\maketitle

\begin{abstract}
The petroleum industry is crucial for modern society, but the production process is complex and risky.
During the production, accidents or failures, resulting from undesired production events, can cause severe environmental and economic damage.
Previous studies have investigated machine learning (ML) methods for undesired event detection. However, the prediction of event probability in real-time was insufficiently addressed,
which is essential since it is important to undertake early intervention when an event is expected to happen.
This paper proposes two ML approaches, random forests and temporal convolutional networks, to detect undesired events in real-time. 
Results show that our approaches can effectively classify event types and predict the probability of their appearance, addressing the challenges uncovered in previous studies and providing a more effective solution for failure event management during the production.\looseness=-1

\end{abstract}

\section{Introduction}
\label{sec:intro}
\vspace{-1ex}

\textbf{Background.}
Petroleum is dubbed as the ``blood'' of modern industry, as it is essential for a wide range of industries. 
The growing awareness of preserving a green planet for our future generations has prompted the petroleum industry to produce energy in a more sustainable practice~\cite{pwc}.
However, the production of petroleum is still a complex and risky process that can have significant adverse consequences, if not managed effectively.
During the petroleum production, accidents or failures, often resulting from undesired events, can cause severe environmental damage. 
For example, oil spills can lead to water pollution and habitat destruction, which will have long-term negative ecological impacts on our society and lead to economic losses.
Therefore, it is vital to detect undesired events during production to minimise environmental damage and protect ecosystems.
Additionally, detecting undesired events can assist engineers in performing accurate failure event management, which will optimise production processes, increase production efficiency, minimise energy consumption, and reduce  maintenance costs. 
A safe and environment-sustainable production will help the petroleum industry to demonstrate social responsibility and addresses the concerns about ecological stewardship.\looseness=-1




\smallskip
\noindent
\textbf{Related Work.}
The petroleum industry has adeptly incorporated AI into the production  through the adoption of digitisation and Industry 4.0 methods.
Due to the large data volume collected from oil well sensors, using AI techniques to assist undesired event detection is becoming feasible.
Some studies~\cite{marins2021RF,turan2021classification,carvalho2021flow,gatta2022predictive,aslam2022anomaly,machado2022improving} have investigated several machine learning (ML) methods for event classification on datasets such as 3W dataset~\cite{vargas2019realistic}. \looseness=-1

\begin{figure}[t]
 \vspace{-1ex}
    \centering
    \includegraphics[width=\textwidth]{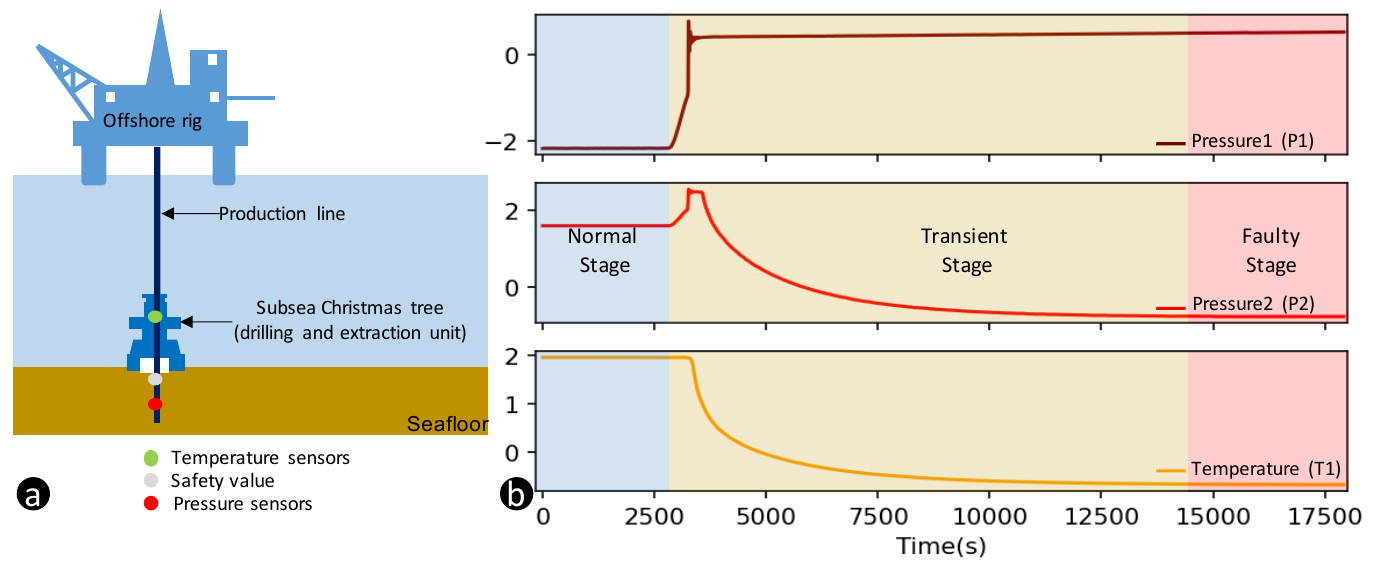}
    \vspace{-1ex}
    \caption{(a) An offshore oil platform and production well (b) Undesired event detection: finding faulty stages via analysing sensor measurements. The illustrated fault is \textit{Spurious downhole safety valve closure}, which means that the system reports that a safety valve has closed, while actually it has not.
    The subjective label provided by the domain expert indicates the faulty stage starts around 15000s, but the actual fault should already start somewhere before 15000s.
    \looseness=-1}
    \label{fig:eventdetect}
    \vspace{-1ex}
\end{figure}

\smallskip
\noindent
\textbf{Challenge.}
However, the previous studies have deficiencies
in detecting and predicting failures.
There are two \textit{key challenges} (C) that should be addressed. 
(C1) the current classification and prediction methods are not performed in real-time, with large window sizes ranging from many minutes to more than one hour. This is insufficient for industrial needs.
(C2) simple classification of faulty stages cannot give accurate information to engineers for when to take intervention:
Although the event detection appears to be a classification task according to the labels in the dataset, it is not always so straightforward, because the labels are subjectively given by domain experts~\cite{vargas2019realistic}. 
From Fig.~\ref{fig:eventdetect}b, it can be seen that some values in the transient stage between normal and faulty stages do not necessarily show drastic changes (Fig.~\ref{fig:eventdetect}): they are not normal but also not fully ``faulty''. 
There is no sharp separation between the transient stage and the faulty stage, which makes conventional classification not accurate enough.
We argue that it is better to predict the probability of emerging failures, rather than simply classifying the time period as normal, transient or faulty.
For having a sustainable production, it is both critical (1) to take early intervention to prevent potential undesired events, and (2) to balance the intervention cost and failure cost to avoid excessive expenditure on false positives. \looseness=-1

\smallskip
\noindent
\textbf{Our Contributions.}
In this paper, to support the sustainable petroleum production, we propose two ML approaches (Sect.~\ref{sec:data}), based on Random Forests (RF) and Temporal Convolutional Networks (TCN), for \textit{real-time} \textit{probabilistic} detection of undesired events. Our approaches can classify the event type and also predict the probability that the given event type appears. The prediction is done for every minute, which we consider a sufficient window length for real-time industry applications. 
Both real-time prediction and probabilistic prediction have been limitedly discussed in past works; 
we are the first (to our best knowledge) to experiment with TCN for the undesired event detection task in the domain.
The evaluation  (Sect.~\ref{sec:eva}) shows very promising results.



\section{Data and Methodology}
\label{sec:data}
\vspace{-1ex}

\noindent 
\textbf{Data Description.}
We investigate the problem with the 3W dataset provided by Petrobras~\cite{vargas2019realistic}, which contains more than 20000 subsets of time series labelled with undesired events, amounting to in total 829,161 minutes (Table~\ref{tab:datastats}).
The dataset consists of both data from real production platforms and simulated data generated from the OLGA system~\cite{SLB}, which is an established tool in the petroleum domain
for providing physically-realistic data.
The data have labels of normal operation and eight undesired event classes (Event1-8). These undesired events are important events that can cause potential accidents or failures (for details, please refer to~\cite{vargas2019realistic}). 
The data are acquired from eight different sensors per second, however, most subsets contain only five features, which are sensor measurements of downhole gauge pressure (P1), transducer temperature (T1) and pressure (P2), upstream choke pressure (P3), and downstream choke temperature (T2). 
The target value, in general, contains normal, transient, and faulty three stages in chronological order (Fig.~\ref{fig:eventdetect}b): the normal stage is always followed by the transient stage. \looseness=-1

\begin{table}[t]
 \vspace{-1ex}
    \centering
    \caption{Statistics of the 3W dataset~\cite{vargas2019realistic}, including normal operation and 8 types of undesired events. The data number is counted by minutes. }
    \label{tab:datastats}
     \vspace{-1ex}
    \setlength{\tabcolsep}{1mm}
    \resizebox{.8\textwidth}{!}{
    \begin{tabular}{c|c|c|c|c}
    \hline
\small
No. & Event Name & \#Real data  & \#Simulated data  & \#Total data  \\ \hline
Normal & Normal operation & 165860 & - & 165860 \\
Event1 & Abrupt basic sediment water increase & 2177 & 150485 & 152602 \\
Event2 & Spurious downhole safety valve closure & 2778 & 7991 & 10769 \\
Event3 & Severe slugging & 9769 & 71619 & 81388 \\
Event4 & Flow instability & 40961 & - & 40961 \\
Event5 & Rapid productivity loss & 6215 & 214534 & 220749 \\
Event6 & Quick production choke restriction & 1306 & 96583 & 97889 \\
Event7 & Scaling in production choke & 5262 & 45345 & 50607 \\
Event8 & Hydrate in production line & 2370 & 35966 & 38336 \\ \hline
    \end{tabular}}
    \vspace{+1ex}
    \vspace{-1ex}
\end{table}

\begin{figure}[t]
 \vspace{-1ex}
    \centering
    \includegraphics[width=.8\textwidth]{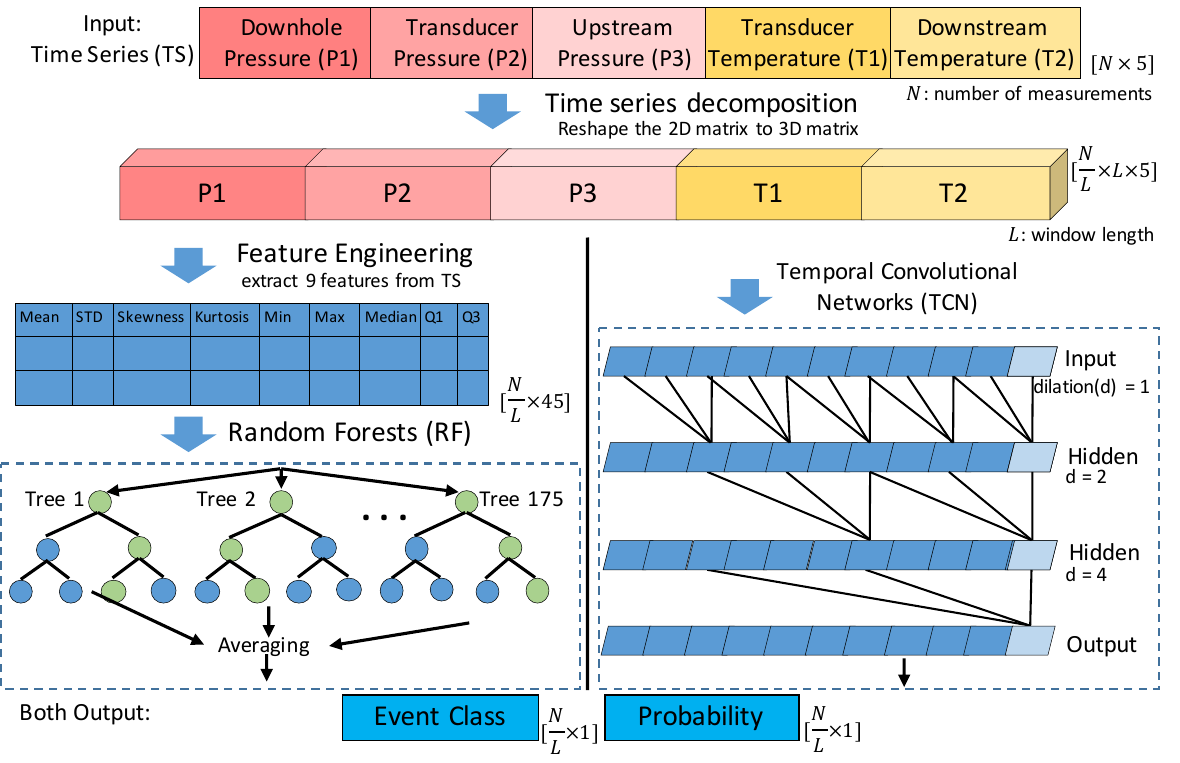}
    \vspace{-1ex}
    \caption{We employ both RF and TCN with the illustrated workflow/architecture for event classification and probability regression, resulting in two RF models and two TCN models for each event, one of each for classification, and another each for regression. We adopt the window length so that  we achieving real-time prediction (every minute) of event class and its probability.
    }
    \label{fig:method}
    \vspace{-1ex}
\end{figure}

\smallskip
\noindent 
\textbf{Methodology.}
Our methods are depicted as the data pipelines in Fig.\ref{fig:method}. First, we select the meaningful features (non-empty, non-constant) as input data, then segment the input data (2D matrices) per minute and reshape them to 3D matrices. After that, the data is fed to 
(1) feature engineering (FE) and random forests and 
(2) to temporal convolutional networks for both event type classification and event probability regression. We choose feature engineering and RF because they are frequently used classic ML methods and have proven to be effective in past works~\cite{marins2021RF}, while TCN has been known as deep learning methods suitable for processing time series~\cite{he2019temporal}. \looseness=-1 

\smallskip
\noindent  \textit{Time Series Decomposition:}
For the purpose of real-time prediction in every minute, the segmentation window length is set to 60, as raw data are collected each second. 

\smallskip
\noindent  \textit{Probability Interpolation:}
The 3W dataset does not provide event probability. To allow probability prediction, we set the probability for normal stages 0, for faulty stages 1, and perform 
linear interpolation for the transient stages.
This approach takes a naive assumption that the probability increases steadily as the production proceeds from normal stages to faulty stages. Although simplistic, it can already provide valuable insights. 
\looseness=-1

\smallskip
\noindent  \textit{Feature Engineering and RF:}
Before feeding the data to the RF model, we perform feature engineering to extract statistical measurements from the input data features. The extracted features are based on the work of ~\cite{marins2021RF}, including mean value, standard deviation, skewness, kurtosis, minimum, maximum, median, and first- and third-quartile values in a window size of 60. 
The extracted features were normalised before using. 

\smallskip
\noindent  \textit{TCN:}
For the TCN model, there is no need for feature engineering, and the reshaped input data can feed directly to the model. Compared with ordinary convolutional neural networks that look at data both in history and the future two directions, the architecture of TCN uses causal convolutional layers to forces the model only look back to the history data to make a prediction, which fits the study task. The filter decides the receptive field of the network, and dilation (d) refers to the spacing between the values in the filter to increase the receptive field.



\smallskip
\noindent 
\textbf{Benefits.}
In contrast to conventional classification approaches in previous works~\cite{marins2021RF,turan2021classification,gatta2022predictive}, our methodology offers two significant benefits. 
Firstly, the prediction window size has been reduced to 60 seconds. We recognise that continuous prediction every second is not always practical or feasible in the industry, as it may consume excessive time and resources and reduce prediction accuracy. 
We aim to achieve a more efficient and accurate prediction than past works~\cite{marins2021RF,turan2021classification,gatta2022predictive} by predicting the event every minute. 
Secondly, our prediction method is based on a probabilistic approach, which provides a prediction of the likelihood of an undesired faulty event. 
This allows domain experts to make informed decisions about when to intervene and take preventive measures. 
The ability to make such decisions in advance can significantly reduce the potential damage caused by a faulty event, and reduce the economic and environmental impact of such incidents.\looseness=-1

\section{Evaluation and Conclusion}
\label{sec:eva}
\vspace{-1ex}

\noindent
\textbf{Experiment Setting and Implementation.}
We experiment with two RF and two TCN models for each event, with one RF and one TCN models for event classification and the other two for event probability regression.
We choose separate models for each event than one single model for all events because of the uneven distribution of event data numbers, as evident from Table \ref{tab:datastats}. 
Previous studies have tested a single classifier for all event classes, resulting in suboptimal results for some events (Event 2, 6, 7)~\cite{marins2021RF,turan2021classification}. 
For example, Event 1 has fifteen times as many data points as Event 2. 
This issue is critical for the industry, because the prediction for each event is essential, and inaccurate predictions can lead to severe consequences. 

The proposed approach is then trained and tested with data from feature engineering for RF. Following common practice, we split 80\% data to training set and 20\% to test set. At the same time, the segmented data for TCN is divided into train-validation (for hyper-parameter tuning)-test data at 70\%-10\%-20\%.
We experiment with various hyper-parameter settings, adopting random search and partial grid search~\cite{turan2021classification}. For the RF, a tree number of 175 and a maximum tree depth of 10 shows the best result. 
For the TCN model in this study, we have set one stack of the residual block with a filter size of 3, dilations [1,2,4], and epochs 30 to get the best result. The Adam algorithm is selected as the optimizer for the TCN model.

We choose common metrics for binary classification, including precision, recall and F1 score, because they consider false positives and false negatives in uneven class distribution, providing more information than accuracy~\cite{marins2021RF}.
We choose root mean squared error (rmse) and mean absolute error (mae) for regression, as rmse is good to measure the differences between target and prediction, and mae is less sensitive to outliers.

\begin{table}[t]
 \vspace{-1ex}
   \centering
\small
\setlength{\tabcolsep}{.5mm}
    \caption{Results for classification (precision, recall, F1 score) and regression (rmse, mae) for the test set. The better results in comparison are  underlined. M1 and M2 stand for our proposed methods Probability-RF and Probability-TCN, respectively. B: baseline of decision trees~\cite{turan2021classification}). }
    \label{tab:result}
\resizebox{1\textwidth}{!}{
\begin{tabular}{c|c c c| c c c| c c c| c c c|c c c|c c c| c c c|c c c}
\toprule
 & \multicolumn{3}{c|}{Event1}  & \multicolumn{3}{c|}{Event2}  & \multicolumn{3}{c|}{Event3}  & \multicolumn{3}{c|}{Event4}   & \multicolumn{3}{c|}{Event5}  & \multicolumn{3}{c|}{Event6}   & \multicolumn{3}{c|}{Event7}  & \multicolumn{3}{c}{Event8}   \\ \midrule
Method  &  M1  &  M2  &  B &  M1  &  M2  &  B &  M1  &  M2  &  B &  M1  &  M2  &  B &  M1  &  M2  &  B &  M1  &  M2  &  B &  M1  &  M2  &  B &  M1  &  M2  &  B \\ \hline
Precision  & \underline{0.95} & 0.80 & 0.83 & \underline{0.98} & \underline{0.98} & 0.42 & 0.99 & \underline{1} & 0.97 & \underline{1} & \underline{1} & 0.94 & \underline{0.97} & 0.90 & 0.91 & \underline{0.97} & 0.81 & 0.76 & \underline{1} & 0.95 &  -  & \underline{0.98} & 0.73 &  0.84 \\
Recall  & 0.95 & 0.80 & \underline{0.98} & \underline{0.98} & \underline{0.98} & 0.60 & 0.99 & \underline{1} & 0.91 & \underline{1} & \underline{1} & 0.96 & \underline{0.97} & 0.90 & 0.94 & \underline{0.97} & 0.81 & 0.87 & \underline{1} & 0.95 &  -  & \underline{0.98} & 0.73 & 0.90 \\
F1 Score  & \underline{0.95} & 0.80 & 0.90 & \underline{0.98} & \underline{0.98} & 0.49 & 0.99 & \underline{1} & 0.94 & \underline{1} & \underline{1} & 0.95 & \underline{0.97} & 0.90 & 0.92 & \underline{0.97} & 0.81 & 0.81 & \underline{1} & 0.95 &  -  & \underline{0.98} & 0.73 & 0.89 \\ \midrule
rmse  & \underline{0.10} & 0.15 &  -  & \underline{0.02} & 0.05 &  -  & \underline{0.00} & \underline{0.00} &  -  & \underline{0.00} & 0.01 &  -  & \underline{0.11} & 0.23 &  -  & \underline{0.09} & 0.24 &  -  & \underline{0.08} & 0.20 &  -  & \underline{0.06} & 0.18 &  -\\
mae  & \underline{0.06} & 0.10 &  -  & \underline{0.01} & 0.02 &  -  & \underline{0.00} & \underline{0.00} &  -  & \underline{0.00} & 0.01 &  -  & \underline{0.04} & 0.14 &  -  & \underline{0.04} & 0.15 &  -  & \underline{0.04} & 0.12 &  -  & \underline{0.04} & 0.15 &  - \\ 
 \bottomrule
\end{tabular}}
\vspace{-2ex}    
\end{table}


\smallskip
\noindent
\textbf{Results and Discussion.}
According to our experiment results in Table \ref{tab:result}, both random forests and temporal convolutional networks models have achieved promising results for the domain users, on the dataset in real-time classification and regression tasks.
The RF model yields superior results to the TCN model (Event 1, 2, 5, 6, 7, 8), while TCN models show better results in Event 3 and 4. This is because Event 3 and 4 have only faulty targets and no normal or transient targets.
Our experimental reproduction indicates that both real-time classification and probabilistic regression models generate good results, while the RF regression model yields the best results.

\begin{figure}[t]
 \vspace{-1ex}
    \centering
    \includegraphics[width=.9\textwidth]{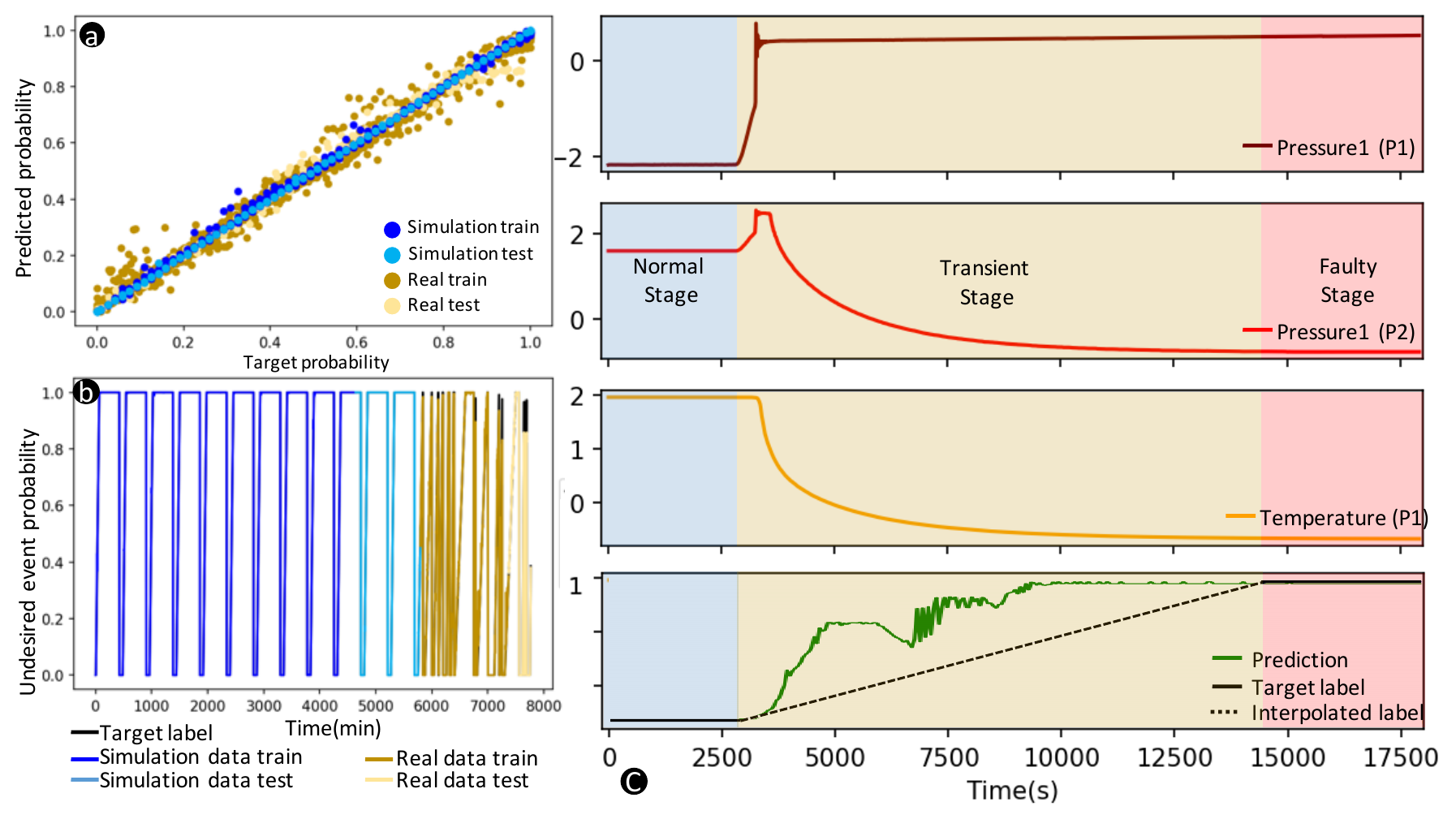}
    \vspace{-1ex}
    \caption{Plots of probabilistic prediction of the \textit{Spurious downhole safety valve closure} events by Probability-RF (a) scatter plot of prediction result and target. (b) line-plot of the no-shuffle data (partial). (c) an example Event2 (Fig.~\ref{fig:eventdetect}) stacked with the target labels and probabilistic prediction.
     }
    \label{fig:prediction}
    \vspace{-1ex}
\end{figure}

Figure \ref{fig:prediction} shows the probabilistic predictions of simulation and real train-test data made by the RF regression model for the undesired event 2. 
The predicted probability results are compared to the target probability labels in Figure 3a, which show impressive good performance.
In Figure 3b, we plot the probability prediction using the 80\%-20\% train-test data split for both simulation and real data. The results show  that the predictions almost overlap completely with the labels for simulation data. For real data, the predictions are also very close to the target labels. 
A comparison of the prediction results for simulation and real data indicates that the latter has lower accuracy. 
This is likely due to real data are more noisy than the simulation data. 
An example prediction of Event 2 (Fig.~\ref{fig:prediction}c) shows that our approach can indeed address challenge 2 (C2) deficiency of simple classification.
Although the labels provided in the dataset show the faulty stage starts at 15000s, our approach can regardless detect that the probability of Event2 is already very high at around 10000s, which should be the actual case judging from the sensor data. 
Providing a probability of undesired events during the transient stage is crucial for the industry, as undesired events are typically labelled based on unfortunate consequences. With probabilistic prediction, early detection can perform more accurately to help avoid excessive expenditure on false positives and minimise the risk of actual undesired events occurring, which will reduce the environmental damage during the production. We also see that the probability prediction is not perfect as the second rise of predicted probability corresponds to no obvious sensor data change. This is due to the limitation of the simplistic interpolation strategy. A better prediction requires more sophisticated interpolation strategy.

\smallskip
\noindent
\textbf{Conclusion.}
This paper proposes ML approaches with random forests and temporal convolutional networks for real-time undesired event detection during petroleum production, which can correctly classify the event type and provide a promising prediction of event probability.
Our work contributes to a more sustainable petroleum production, by predicting event probability to engineers for performing event management and timely intervention to prevent undesired events. 
This proactive approach could help the industry to minimise ecological impacts, increase production efficiencies, reduce maintenance costs, and mitigate the growing concerns about industrial sustainable development.
In future research, we plan to improve the probability prediction with better interpolation strategy and study the role of the work in the context of the digital twins.

\medskip

\noindent \textbf{Acknowledgements} This work is supported by the Norwegian Research Council via PeTWIN (294600), DigiWell(308817) and SIRIUS (237898).

\begin{footnotesize}

\bibliographystyle{unsrt}
\bibliography{PRICAI2023}

\end{footnotesize}

\end{document}